\pdfoutput=1

\documentclass[11pt]{article}

\usepackage[review]{acl}

\usepackage{times}
\usepackage{latexsym}
\usepackage{colortbl}
\usepackage{longtable}

\usepackage[T1]{fontenc}
\usepackage{hyperref}
\usepackage{booktabs}
\usepackage{multirow}
\usepackage{multicol}
\usepackage{makecell}
\usepackage{graphicx}

\usepackage[utf8]{inputenc}

\usepackage{microtype}

\title{LLMs Know What They Need: Leveraging a Missing Information Guided Framework to Empower Retrieval-Augmented Generation}

    \author{Keheng Wang, \ {\bf Feiyu Duan}, \ {\bf Peiguang Li}, \ {\bf Sirui Wang}, \ {\bf Xunliang Cai} \\ \\
    \{wangkeheng, duanfeiyu, lipeiguang, wangsirui, caixunliang\}@meituan.com}

\begin{document}
\maketitle
\begin{abstract}
Retrieval-Augmented Generation (RAG) demonstrates great value in alleviating outdated knowledge or hallucination by supplying LLMs with updated and relevant knowledge. However, there are still several difficulties for RAG in understanding complex multi-hop query and retrieving relevant documents, which require LLMs to perform reasoning and retrieve step by step. Inspired by human's reasoning process in which they gradually search for the required information, it is natural to ask whether the LLMs could notice the missing information in each reasoning step. In this work, we first experimentally verified the ability of LLMs to extract information as well as to know the missing. Based on the above discovery, we propose a Missing Information Guided Retrieve-Extraction-Solving paradigm (MIGRES), where we leverage the identification of missing information to generate a targeted query that steers the subsequent knowledge retrieval. Besides, we design a sentence-level re-ranking filtering approach to filter the irrelevant content out from document, along with the information extraction capability of LLMs to extract useful information from cleaned-up documents, which in turn to bolster the overall efficacy of RAG. Extensive experiments conducted on multiple public datasets reveal the superiority of the proposed MIGRES method, and analytical experiments demonstrate the effectiveness of our proposed modules. Code and data are released in \href{https://github.com/AdelWang/MIGRES}{https://github.com/AdelWang/MIGRES}.
\end{abstract}

\section{Introduction}

Large Language Models (LLMs) have recently shown impressive capabilities across a wide range of Natural Language Processing (NLP) tasks~\cite{ouyang2022training, touvron2023llama, openai2023gpt4}. Nevertheless, LLMs only possess the knowledge present in their pretraining stage and could not remember them completely, hence LLMs may fail to answer or prone to generate hallucinations given the questions that beyond their knowledge scope.~\cite{bang2023multitask, huang2023survey}. 

Retrieval-Augmented Generation (RAG) is a promising solution to improve the accuracy of responses~\cite{Khandelwal2020Generalization, izacard2022atlas}, which adopts a retrieve-then-generate setup, i.e., it first retrieves query-related documents from external corpus and then request LLMs generates responses conditioning on the knowledge in these documents. Despite the effectiveness, RAG still faces several challenges and we classify them into doc-related and query-related: 

For query side, the user input query might be complex and multi-hop (e.g., \emph{What is the place of birth of the director of film Oh Billy, Behave?}), where the required information (\emph{the name of the director of film Oh Billy, Behave}) may not be explicitly stated in the query, making it difficult to retrieve relevant documents \cite{shao-etal-2023-enhancing}. For document side, retrieving relevant documents from the extensive candidates is inherently challenging \cite{gao2024retrievalaugmented, sun2024verifiable}, and moreover, there often exist irrelevant noise content throughout the complete document.

Chain-of-Thought (CoT)~\cite{wei2022chain} is introduced in RAG to tackle complex multi-hop issues by breaking them down into couples of single-hop tasks. However, traditional CoT-based methods are susceptible to hallucination generation during the query decomposition and they often require the integration of task-specific demonstrations to improve the reasoning quality. Inspired by human's reasoning process in which they gradually search for the required information, it is natural to ask whether the LLMs could notice the missing information in each reasoning step. 

To answer this question, we first examined if LLMs could identify what information is missing to answer the query conditioning on providing partial knowledge as known part. Surprisingly, we discover that, even in zero-shot scenarios, LLMs can precisely identify what knowledge points is missing with an average accuracy of 95.6\% (More details of this experiments are discussed in Section 2).

Motivated by this discovery, we propose a Missing Information Guided Retrieve-Extraction-Solving paradigm (MIGRES). In query side, we further prompt LLMs using the missing information to formulate straightforward single-hop queries. These new single-hop queries can guide the subsequent knowledge retrieval process, thus improving the performance when dealing with complex multi-hop challenges. 

As for the document side challenges, we perform a similar attempt, and our experiments indicate that LLMs are adept at extracting useful information from denoised documents. Therefore, in order to filter out the irrelevant content and provide LLMs with denoised documents, we introduce a sentence-level re-ranking filtering strategy. This method breaks down the retrieved documents into sentences and computes a relevance score for each sentence, which allows for their re-ranking and filteration.

\section{Preliminary Experiments}
\label{preliminary}
In recent studies on RAG, LLMs have been commonly used to summarize documents and extract information \cite{gao-etal-2023-enabling, sun2024verifiable}. Previous works also addressed complex multi-hop queries by decomposing them into sub-questions using LLMs \cite{self-ask, yao2023react, wang2023knowledgedriven}, yielding promising results and demonstrating the forward-looking ability of LLMs. In this section, we aim to (1) further investigate the ability of LLMs to effectively extract accurate knowledge from retrieved documents; and (2) explore whether the model can infer the remaining information needed to solve a query based on the known information. We conduct experiments on the 2WikiMultiHop \cite{xanh2020_2wikimultihop} and the Musique \cite{trivedi2021musique} datasets. These two datasets provide intermediate supervised signals as evidence and annotate the corresponding documents, which facilitate our investigation.
\subsection{Settings}
\textbf{Pre-process}
We randomly sample 500 instances from the training set, then prompt LLMs to generate the intermediate QA pairs given the original question, evidence, and the final answer. 
Examples can be found in Figure \ref{prompt_decom}.

\noindent \textbf{Information Extraction} We utilize BM25 to search top 50 relevant passages from the external corpus, and merge them with the original passages provided in these two datasets. A passage is labeled positive if it contains the sub-answer in the evidence\footnote{We utilize the evaluation code in \citet{karpukhin2020dense}}. Then we randomly sample 5 passages and concatenate them with the original question as well as the decomposed sub-questions, and prompt the LLM to extract useful information from the passages and cite them accordingly \cite{gao-etal-2023-enabling}. 

\noindent \textbf{Missing Information Generation} We randomly concatenate all or partial information in the QA pairs obtained in the pre-process step, then prompt the LLM to determine whether the question can be answered, and generate the missing information accordingly. We then evaluate if the generated missing information aligns with subsequent sub-questions.


We utilize gpt-3.5-turbo as the backend LLM. All experiments are conducted under a zero-shot setting. All prompts and cases can be found in Appendix \ref{appendix_prompt_prelimary}.


\subsection{Results}
The experimental results are shown in Table \ref{ie_res} and Table \ref{mg_res}. 
ChatGPT demonstrated commendable performance in the area of information extraction, achieving precision scores of 89.3 on WikiHop and 91.8 on Musique, along with a 72.0 and 76.1 recall scores. 
\begin{table}[t]
    \centering
    \scalebox{0.85}{
    \begin{tabular}{c|cc|cc|c}
    \toprule
    \textbf{Dataset} & \textbf{Prec.} & \textbf{Rec.} & \textbf{Prec.$^\dag$} & \textbf{Rec.$^\dag$} & \textbf{Useful} \\
     \hline
     Wikihop & 89.3 & 72.0 & 94.8 & 72.0 & 96.6 \\
     Musique & 91.8 & 76.1 & 94.6 & 74.2 & 94.3 \\
     \bottomrule
    \end{tabular}}
    \caption{Experimental results on information extraction. \textbf{Entail} and \textbf{Useful} are scored with binary classification accuracy. \textbf{Prec.$^\dag$} and Rec.$^\dag$ denote the precision and recall after entailment judgement.}
    \label{ie_res}

    \scalebox{0.85}{
    \begin{tabular}{c|cc|cc}
    \toprule
    {} & \multicolumn{2}{c|}{\textbf{All}} & \multicolumn{2}{c}{\textbf{Partial}} \\
    {\textbf{Dataset}} & \textbf{Acc} & \textbf{Match} & \textbf{Acc} & \textbf{Match} \\
     \hline
     Wikihop & 91.2 & - & 98.4 & 98.0 \\
     Musique & 86.8 & - & 92.8 & 96.8 \\
     \bottomrule
    \end{tabular}}
    \caption{Experimental results on missing information generation. \textbf{All} signifies the integration of all the information in decomposed QA pairs, whereas \textbf{Partial} concatenates partial or no information. \textbf{Match} denotes the accuracy that the missing information aligns with the subsequent decomposed sub-questions.}
    \label{mg_res}
\end{table}

\begin{figure*}[t]
    \centering
    \includegraphics[width=0.9\linewidth]{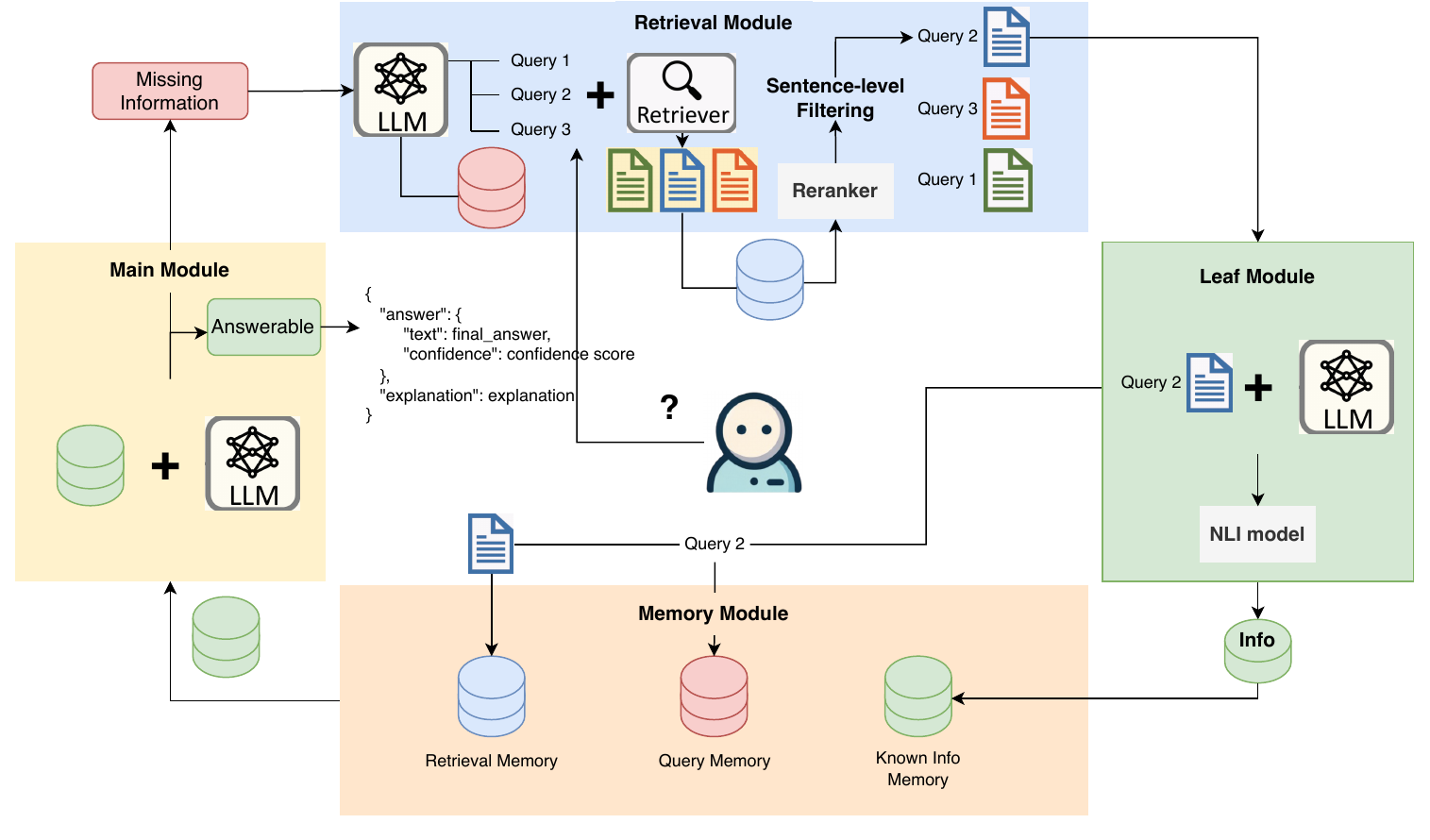}
    \caption{The overall framework of MIGRES.}
    \label{method}
\end{figure*}



As for missing information generation, when all necessary information is provided, the LLM achieves an accuracy of 91.2\% and 86.8\%, where it outputs a precise answer.
In scenarios where the available knowledge is incomplete, the LLM demonstrates a notable accuracy of 95.6\% on average, and can generate missing information that aligns with the subsequent decomposed sub-questions.
This highlights the model's proficiency in identifying what additional information is necessary to resolve the question and in generating the missing information that can be used to guide the subsequent knowledge retrieval.

\section{Methodology}
Inspired by the discovery in Section \ref{preliminary}, we propose MIGRES, a Missing Information Guided Retrieve-Extraction-Solving paradigm that leverages the identification of missing information to guide the subsequent knowledge retrieval, and utilizes the extracted concise useful information for solving factoid question-answering tasks. In this section, we present in detail the framework of our proposed method.

\subsection{Overview}
\label{overall framework}
As shown in Figure \ref{method}, MIGRES contains the following flowline modules:
\begin{itemize}
    \item \textbf{Main module}, which takes the query and useful information extracted from the retrieved contexts as input to determines whether the question can be answered. It will return the final answer with an explanation or the missing information depending on its determination.
    \item \textbf{Retrieval module}, which is consists of a Query Generator, a Retriever and a Knowledge Filter. Given the original query, previously asked queries, useful information extracted, and the missing information output by \textbf{Main module}, the Query Generator will first generate simpler and diverse new queries to facilitate the subsequent retrieval. Then the Retriever obtains relevant external knowledge in response to these queries, which undergo filteration through Knowledge Filter to remove noise at the level of passages and sentences.
    \item \textbf{Leaf module}, which reads the retrieved external knowledge to extract useful information with citation of support passages. Considering that the extracted information might include hallucinations, we incorporate a subsequent evaluation step to ascertain if the cited passages indeed entail the information extracted through an NLI model.
    \item \textbf{Memory module}, which is used to record the historical retrieved passages and the generated queries.
\end{itemize}

Upon receiving a question input, MIGRES begins with knowledge retrieval, initially employing a Retriever to obtain pertinent passages from external knowledge, followed by the Leaf Module for the useful information distillation. The distilled information, together with the original question, is then fed into the Main Module, which will assess if the current information suffices to answer the question and provides a response accordingly. Should the available information prove insufficient, the Main Module identifies the missing information, which then serve to generate new queries for subsequent knowledge retrieval. This process is iteratively repeated until either a definitive answer is given or a predetermined max iteration steps $\mathcal{T}$ is reached. Prompts and cases for each Module can be found in Table \ref{migres_case}.

\subsection{Detailed introduction of each Module}
\noindent \textbf{Main Module} 
The Main Module is comprised of an LLM tasked with determing whether the input query can be solved based solely on the known information. If the information is sufficient to conclude an answer, the LLM will generate a concise answer with an explanation. Otherwise, it will output "unanswerable" and identify what information is lacking. The missing information is then fed into the Retrieval Module for new queries generation and subsequent knowledge retrieval.


\noindent \textbf{Retrieval Module} Within the Retrieval Module, we instruct a LLM, which serves as the Query Generator, to formulate no more than three new distinct queries based on the missing information, and utilize them to retrieve external knowledge. A re-rank model is then utilized to calculate a relevance score between external knowledge and queries, and passages with relevance lower than a threshold $\theta$ are directly filtered out. However, the remained passages might 1) contain sentence-level noise, as only a small portion of knowledge within the retrieved passages is useful in most cases; and 2) be empty, which implies that external knowledge may not encompass information pertinent to the queries. We thus propose the following two strategies to address these two issues:
\begin{itemize}
    \item \textbf{Sentence-Level Re-rank and Filtering} We utilize NLTK to segment the passage into individual sentences, and compute a relevance score for each sentence using the same Re-rank model. The noisy sentences with relevance lower than $\theta$ are then filtered out. Should the relevance of the entire passage surpass that of all the individual sentences, we opt to preserve the original passage as the ultimate knowledge.
    \item \textbf{LLM Knowledge prompting} While it's possible that an ineffective query could result in subpar retrieval or relevance, to improve the iteration efficiency of MIGRES when pertinent external knowledge is lacking, and to fully utilize the LLM's parametric knowledge, we prompt the LLM to generate relevant information in response to the input queries when no remaining passage is available, treating it as the ultimate knowledge.
\end{itemize}

\noindent \textbf{Leaf Module} Subsequently, both the queries and the acquired knowledge are fed into the Leaf Module to obtain distilled useful information. Inspired by previous work \cite{gao-etal-2023-enabling, sun2024verifiable}, we also instruct the LLM to concurrently cite the indices of the passages that substantiate the extracted information. This practice proves beneficial in diminishing instances of hallucination and eradicating unfounded generated content. To further avoid obtaining hallucinated information, we utilize an NLI model to determine whether the cited passages indeed entail the distilled information, and filter out that is not entailed.

\noindent \textbf{Memory Module} We observe that when the Leaf Module fails to recall useful information, the Query Generator tends to produce queries identical to previous ones. Additionally, the retrieved knowledge may include hard negative passages that scored high in relevance to the query yet fail to provide useful information, continually incorporating such knowledge does not contribute to resolving the question.
To address the above issues, we propose to utilize a Memory Module to track the queries generated and the external knowledge retrieved, and thus improve the diversity of new queries and filter out the hard negative knowledge.

\section{Experiment}

\subsection{Experiment setup}
We conduct experiments on five datasets across three diverse knowledge-intensive tasks under a zero-shot setting: (1) \textbf{Multi-hop question answering}, including 2WikiMultihopQA (Wikihop) \cite{xanh2020_2wikimultihop}, HotpotQA \cite{yang-etal-2018-hotpotqa} and Musique \cite{trivedi2021musique}; (2) \textbf{Open-domain question answering}, we use Natural Question \cite{kwiatkowski-etal-2019-natural} and TriviaQA \cite{joshi-etal-2017-triviaqa}; (3) \textbf{Commonsense reasoning}, which includes StrategyQA \cite{geva2021strategyqa}. 

We utilize gpt-3.5-1106 \footnote{\href{https://openai.com/}{https://openai.com}} as our backend LLM within all modules, and utilize BM25 with $k1=0.9$ and $b=0.4$ as the Retriever, the BGE-reranker-base \cite{bge_embedding} as the Re-rank model, and T5-xxl-nli \cite{honovich-etal-2022-true-evaluating}\footnote{Downloaded from \href{https://huggingface.co/google/t5_xxl_true_nli_mixture}{t5\_xxl\_true\_nli\_mixture}} as the NLI model, respectively. Following previous work \cite{sun2024verifiable}, we conduct evaluations on all 229 questions from StrategyQA and randomly sub-sample 200 questions from all other datasets for saving the cost of running experiments. We compare the final answer output by LLMs with the reference answer using exact match (EM) after normalization. As most experiments are conducted under a zero-shot setting, we also evaluate the correctness of model outputs using gpt-3.5-1106 for more robust evaluation, which is proved to be reliable in \citet{shao-etal-2023-enhancing}. We denote the resulting metric as Acc$^{\dag}$, the prompt is shown in Table \ref{other_prompt}.



We use the December 2017 Wikipedia dump \cite{izacard2022atlas} for HotpotQA, the December 2018 dump \cite{karpukhin2020dense} for Wikihop and ODQA, and the December 2021 Wikipedia dump \cite{izacard2022atlas} for Musique and StrategyQA. To guarantee the retrieval of pertinent knowledge, we also create an oracle version for each dump, in which we augment all contexts in the original development sets of Multi-hop QA and Commonsense QA (including distractors) into the retrieval corpus. The results using oracle dump are denoted as MIGRES$^{\dag}$. All datasets, prompts and hyper-parameters are summarized in Table \ref{hyper-setting}.

\begin{table*}[ht]
    \centering
    \scalebox{0.85}{
        \begin{tabular}{lcccccccccccc}
            \toprule
            \textbf{Datasets} & \multicolumn{2}{c}{\textbf{Wikihop}} & \multicolumn{2}{c}{\textbf{Hotpot}} & \multicolumn{2}{c}{\textbf{Musique}} & \multicolumn{2}{c}{\textbf{NQ}} & \multicolumn{2}{c}{\textbf{TriviaQ}} & \multicolumn{2}{c}{\textbf{StrategyQA}} \\
            \hline
            \textbf{Metrics} & \textbf{EM} & \textbf{Acc$^\dag$} & \textbf{EM} & \textbf{Acc$^\dag$}& \textbf{EM} & \textbf{Acc$^\dag$} & \textbf{EM} & \textbf{Acc$^\dag$} & \textbf{EM} & \textbf{Acc$^\dag$} & \textbf{EM} & \textbf{Acc$^\dag$} \\
            \hline
            \rowcolor{lightgray}\multicolumn{13}{c}{Zero-shot} \\
            \hline
            {VANILLA} & 24.5 & 45.5 & 27.5 & 60.5 & 10.5 & 26.0 & 33.5 & 72.5  & 57.5 & 87.5 & 62.9 & 62.9 \\
            {VANILLA-s} & 26.5 & 48.5  & 28.0 & 62.0 & 11.5 & 27.5 & 34.5 & 75.5 & 59.0 & 88.0 & 65.5 & 65.5  \\
            {SUMM} & 25.5 & 52.0  & 25.5 & 56.0 & 10.0 & 29.5 & 32.5 & 72.5 & 56.5 & 84.5 & 62.9 & 62.9 \\
            {SNIPPET} & 26.0 & 50.5  & 29.5 & 61.5 & 10.5 & 27.5 & 33.0 & 69.0 & 57.5 & 87.5 & 62.0 & 62.0  \\
            {RERANK} & 28.5 & 56.0 & 30.0 & 63.0 & 12.5 & 30.0 & 36.5 & 77.0 & 59.0 & 89.5 & 65.9 & 65.9  \\
            \hline
            {MIGRES} & 33.6 & 58.5 & 38.0 & 66.6 & 18.6 & 32.8 & 43.0 & \textbf{80.0} & \textbf{61.0} & 91.0 & \textbf{73.4} & \textbf{73.4} \\
            {MIGRES$^\dag$} & \textbf{40.1} & \textbf{65.9}  & \textbf{38.6} & \textbf{67.2} & \textbf{19.4} & \textbf{33.2} & \textbf{43.5} & 79.0 & 60.5 & \textbf{91.0} & 72.1 & 72.1 \\
            \hline
            \rowcolor{lightgray} \multicolumn{13}{c}{Few-shot} \\
            \hline
            ITRG & 29.8 & -  & 33.4 & - & - & - & 33.8 & -  & \textbf{77.8} & - & - & - \\
            VTG & 41.5 & -  & - & - & - & - & \textbf{63.0} & -  & - & - & - & - \\
            ReAct & 28.0 & 45.9 & 24.9 & 61.1 & 23.4 & 37.9 & - & - & - & - & 66.9 & 66.9  \\
            Self-Ask & 37.3 & 55.9 & 36.8 & 64.8 & \textbf{27.6} & \textbf{42.9} & - & - & - & - & 70.2 & 70.2  \\
            ITER-RETGEN & 34.9 & 58.1 & 44.1 & 71.2 & 26.4 & 41.0 & - & - & - & - & 73.0 & 73.0  \\
            \hline
            {MIGRES$^*$} & 47.6 & 61.2 & 46.8 & 68.6 & 19.6 & 34.0 & 47.0 & 78.0 & 63.0 & 91.5 & \textbf{74.2} & \textbf{74.2} \\
            {MIGRES$^{*\dag}$} & \textbf{54.0} & \textbf{71.0}  & \textbf{49.4} & \textbf{72.4} & 20.8 & 34.6 & 48.0 & \textbf{80.0} & 62.5 & \textbf{92.0} & 72.9 & 72.9 \\
            \bottomrule
        \end{tabular}}
        \caption{Comparison between MIGRES and baselines on Multi-hop QA, Open-domain QA and Commonsense QA task. Acc$^{\dag}$ is the accuracy of model outputs evaluated with gpt-3.5-1106. MIGRES$^{\dag}$ is the results using the oracle knowledge pool. The results of "Few-shot" are extracted from the original paper, "-" represents that the results are unavailable. The best values are highlighted in \textbf{bold}.}
        \label{main_res}
\end{table*}

\begin{table*}[ht]
    \centering
    \scalebox{0.8}{
        \begin{tabular}{cccccccccc}
            \toprule
            {\textbf{Method}} & \multicolumn{3}{c}{\textbf{Wikihop}} & \multicolumn{3}{c}{\textbf{Hotpot}} & \multicolumn{3}{c}{\textbf{Musique}} \\
            \hline
            {} & \textbf{\# API} & \textbf{\# Iter} & \textbf{\# Passages} & \textbf{\# API} & \textbf{\# Iter} & \textbf{\# Passages}& \textbf{\# API} & \textbf{\# Iter} & \textbf{\# Passages} \\
            \hline
            ReAct & 3.0 & 3.0 & 15.0 & 2.9 & 2.9 & 14.4 & 2.9 & 2.9 & 14.3 \\
            Self-Ask & 3.2 & 3.2 & 15.9 & 3.0 & 3.0 & 14.8 & 3.2 & 3.2 & 16.0 \\
            ITER\_RETGEN & 2.0 & 2.0 & 5.0 & 2.0 & 2.0 & 5.0 & 2.0 & 2.0 & 5.0 \\
            MIGRES & 8.4 & 2.5 & 1.3 & 9.5 & 2.7 & 3.1 & 6.4 & 1.9 & 2.3 \\
            MIGRES$^{*}$ & 8.1 & 2.4 & 1.4 & 11.9 & 3.4 & 1.8 & 5.8 & 1.8 & 2.6 \\
            \hline
            {} & \multicolumn{3}{c}{\textbf{NQ}} & \multicolumn{3}{c}{\textbf{TriviaQ}} & \multicolumn{3}{c}{\textbf{StrategyQA}} \\
            \hline
            {} & \textbf{\# API} & \textbf{\# Iter} & \textbf{\# Passages}& \textbf{\# API} & \textbf{\# Iter} & \textbf{\# Passages}& \textbf{\# API} & \textbf{\# Iter} & \textbf{\# Passages} \\
            \hline
            ReAct & - & - & - & - & - & - & 2.9 & 2.9 & 14.3 \\
            Self-Ask & - & - & - & - & - & - & 3.2 & 3.2 & 16.0 \\
            ITER\_RETGEN & - & - & - & - & - & - & 2.0 & 2.0 & 5.0 \\
            MIGRES & 3.5 & 1.3 & 1.3 & 3.4 & 1.1 & 1.3 & 4.0 & 1.3 & 1.3 \\
            MIGRES$^{*}$ & 4.9 & 1.3 & 1.4 & 4.0 & 1.3 & 1.4 & 4.7 & 1.5 & 1.4 \\
            \bottomrule
        \end{tabular}}
        \caption{Efficiency of MIGRES. \textbf{\# API}, \textbf{\# Iter} and \textbf{\# Passages} represent the average API calls, iteration steps and number of passages within each iteration. It can be seen that the total number of passages is less than 5 across all datasets.}
        \label{efficiency}
\end{table*}

\subsection{Baselines}
We consider the following baselines for comparison.



\noindent \textbf{ALCE} \cite{gao-etal-2023-enabling}, which includes (1) \textit{VANILLA}, where top-k ranked documents are concatenated as knowledge augmentation for prompting LLMs to generate responses. We also evaluate the effectiveness of incorporating sentence-level filtering, denoted as \textit{VANILLA-s} (2) \textit{SUMM / SNIPPET}, where the LLM is required to synthesize relevant information or extract snippets from the top-k ranked documents. This condensed text is then integrated into the prompt for generating the response. (3) \textit{RERANK} This method prompts the LLM to generate four distinct responses and then choose the answer with the highest citation recall as the final output. We evaluate all these methods under a \textbf{zero-shot} setting, and utilize the Re-rank model to re-rank the retrieved knowledge.

\noindent \textbf{ITRG} \cite{feng2023retrievalgeneration}, which is a pipeline that utilize both generation augmented retrieval and retrieval augmented generation, and iteratively retrieve knowledge based on the previous generated content.

\noindent \textbf{VTG} \cite{sun2024verifiable}, which is similar to ALCE while they utilize a verifier to evaluate whether the retrieved knowledge entails the generated sentence. If not, they prompt LLMs to generate new queries for searching more evidence that supports the current sentence and drop any unsupported sentence.

\noindent \textbf{ReAct} \cite{yao2023react} includes reasoning, action and observation steps, where the action can be either generating a query for searching information or conclude an answer, and the observation is to concatenate the retrieved knowledge.

\noindent \textbf{Self-Ask} \cite{self-ask} includes question decomposition and answer searching steps. The LLM gives the final answer until no more follow-up questions are generated. \citet{yoran2023answering} and \citet{shao-etal-2023-enhancing} further prepend newly retrieved knowledge to the original question for sub-answers generation. 

\noindent \textbf{ITER-RETGEN} \cite{shao-etal-2023-enhancing} combines retrieval augmented generation with generation augmented retrieval that iteratively generates new sentences as extensions to the original query for next step retrieval.

As most baselines are under a few-shot setting, we also conduct experiments concatenating demonstrations to perform a 2-shot ICL. We design specific examples randomly sampled from the training set of Wikihop, NQ and StrategyQA for each modules and fix then during evaluation. The same demonstrations are shared in multi-hop QA, ODQA and Commonsense QA, respectively. Our method with few-shot learning is denoted as MIGRES$^{*}$, and if not specified, the default setting for MIGRES is zero-shot.

\begin{table*}[t]
    \centering
    \scalebox{0.8}{
    \begin{tabular}{c|ccccccccc}
        \toprule
        \textbf{Dataset} & \multicolumn{3}{c}{\textbf{Wikihop}} & \multicolumn{3}{c}{\textbf{Musique}} & \multicolumn{3}{c}{\textbf{NQ}} \\
        \hline
        \textbf{Metrics} & \textbf{EM} & \textbf{Acc$^\dag$} & \textbf{\# Tokens} & \textbf{EM} & \textbf{Acc$^\dag$} & \textbf{\# Tokens} & \textbf{EM} & \textbf{Acc$^\dag$} & \textbf{\# Tokens} \\
        \hline
        MIGRES & 33.6 & 58.5 & 733 & 18.0 & 31.5 & 1224 & 43.0 & 80.0 & 333 \\
        {$_{w/o \ Sentence \ Filtering}$} & 33.0 & 58.0 & 877 & 19.0 & 33.6 & 1697 & 43.5 & 76.5 & 404 \\
        {$_{w \ SUMM}$} & 26.0 & 54.2 & 1249 & 12.0 & 30.0 & 1898 & 40.0 & 73.5 & 522 \\
        {$_{w \ SNIPPET}$} & 26.6 & 52.2 & 1095 & 12.5 & 29.5 & 1775 & 42.5 & 74.0 & 454 \\
    \bottomrule
    \end{tabular}}
    \caption{Comparison of MIGRES different knowledge compression/filtering method. \textbf{\# Tokens} denotes the average tokens consumption of external knowledge for each instance in the Leaf Module. For \textit{SUMM} and \textit{SNIPPET}, we also count the token consumption calling API for summarization and snippet extraction.}
    \label{comparison_compress_res}
\end{table*}

\begin{table*}[t]
    \centering
    \scalebox{0.85}{
    \begin{tabular}{c|ccccccccc}
        \toprule
        \textbf{Dataset} & \multicolumn{3}{c}{\textbf{Wikihop}} & \multicolumn{3}{c}{\textbf{Musique}} & \multicolumn{3}{c}{\textbf{NQ}} \\
        \hline
        \textbf{Metrics} & \textbf{EM} & \textbf{Acc$^\dag$} & \textbf{\# Avg. Iter} & \textbf{EM} & \textbf{Acc$^\dag$} & \textbf{\# Avg. Iter} & \textbf{EM} & \textbf{Acc$^\dag$} & \textbf{\# Avg. Iter} \\
        \hline
        GPT-3.5-1106 & 34.8 & 54.4 & 2.53 & 16.8 & 31.6 & 2.96 & 42.5 & 73.5 & 1.28 \\
        {$_{w/o \ GPT \ knowledge}$} & 32.0 & 53.8 & 2.96 & 15.6 & 30.0 & 2.95 & 41.5 & 72.5 & 1.51 \\
        \hline
        GPT-4-0613 & 50.0 & 68.0 & 2.76 & 22.4 & 39.8 & 2.85 & 44.5 & 77.5 & 1.31 \\
        {$_{w/o \ GPT \ knowledge}$} & 46.6 & 63.8 & 3.03 & 21.4 & 41.8 & 2.80 & 45.0 & 75.5 & 1.50 \\
    \bottomrule
    \end{tabular}}
    \caption{Comparison of MIGRES with/without prompting LLM to generate relevant information when no documents retrieved have a relevance score higher than $\delta$ (to avoid that some questions consistently fail to retrieve highly relevant documents, we set $\delta = 1.0$). \textbf{\# Avg. Iter} denotes the mean iteration steps during inference.}
    \label{res_knowledge}
\end{table*}

\subsection{Main results}
As shown in Table \ref{main_res}, MIGRES outperforms all methods in ALCE under the zero-shot setting, and even achieves competitive or better results on Wikihop, HotpotQA and StrategyQA compared with few-shot baselines, demonstrating the effectiveness of our proposed method. The performance of MIGRES can be further improved when augmenting oracle knowledge, indicating that instances of incorrect responses from the LLM are sometimes a result of the absence of relevant knowledge in the external retrieval corpus.

We also design demonstrations to conduct few-shot learning. It can be seen from Table \ref{main_res} that the performance generally improves, as we find that adding demonstrations can steer LLM to generate more targeted new queries, and reduce the hallucinated knowledge generated. The EM scores on the Wikihop and HotpotQA datasets, which feature more standardized answers, saw notable improvement. This implies that few-shot learning effectively aids the model in avoiding the creation of unnecessary descriptions and in delivering more standardized responses.

MIGRES performs poor on Musique, we think it's because that questions in this dataset are more obscure and ambiguous, making it difficult for the Retriever to recall relevant knowledge from retrieval corpus. For example, Natalie Wood and Mara Wilson both played Susan Walker in Miracle on 34th Street, but MIGRES fails to recall the knowledge about Natalie Wood and only output Mara Wilson as the player, while the only label provided for this instance is Natalie Wood, resulting an incorrect response for the question \emph{Who is the sibling of the actress who played Susan Walker in Miracle on 34th street?}.

It's worth noting that, with irrelevant knowledge filtering, we greatly reduce the token consumption of external knowledge, while keeping superior or competitive performance compared with baselines. As shown in Table \ref{efficiency}, despite additional API calls are required in Main Module, Retrieval Module and gpt knowledge prompting, we reduce the total number of passages, which is much more costly, to less than 5.

\subsection{Benefit of Sentence-Level Filtering}
We adopt a sentence-level filtering to further reduce the noise in the retrieved passages. As can be ssen from Table \ref{main_res}, VANILLA-s consistently improves the performance across all datasets compared with VANILLA, and reduces the token cost of external knowledge. Incorporating sentence-level filtering also outperforms SUMM and SNIPPET on various datasets without additional calls of LLM, but its performance is less effective on Wikihop and Musique. This could be attributed to the fact that VANILLA retains only the top 5 passages re-ranked, while content deemed irrelevant is not kept in the SUMM and SNIPPET. With 7.9 and 8.4 average calls of LLM for knowledge compression to get 5 relevant refined passages, they are able to capture more information, thereby improve the performance, especially in multi-hop QA scenarios. A comparison of these three compression methods can be found in Table \ref{compress_res}.

To further investigate the token efficiency, we evaluate MIGRES useing different compression methods on Wikihop, Musique and NQ, the results are shown in Table \ref{comparison_compress_res}. We can see that employing sentence-level filtering slightly outperforms the others, with less token consumption and no extra calls for LLM.

\subsection{Benefit of Prompting GPT knowledge}
The Retrieval Module may sometimes fail to deliver valid pertinent knowledge, either due to the imprecision of the Retriever or the Re-rank model, or because the corresponding external knowledge is absent. Additionally, the LLM possesses a wealth of world knowledge that enables it to generate valuable information given the query. To fully leverage the internal knowledge of LLM and to improve the iteration efficiency of MIGRES, we prompt the LLM to generate query-related information when the Retrieval Module returns no knowledge. 

\begin{table}[h]
    \centering
    \scalebox{0.9}{
    \begin{tabular}{c|ccc}
    \toprule
    \textbf{Dataset} & \textbf{Wikihop} & \textbf{Musique} & \textbf{NQ} \\
    \hline
    gpt-3.5-1106 & 45.5 & 54.6 & 24.7 \\
    gpt-4-0613  & 87.7 & 77.5 & 32.0 \\
    \hline
    \end{tabular}}
    \caption{Accuracy of the generated knowledge.}
    \label{acc_knowledge}
\end{table}

To investigate the effectiveness of utilizing the internal knowledge of LLM, we conduct experiments without knowledge prompting, where MIGRES attempts to retrieve relevant knowledge at the next iteration steps by generating more simper and diverse queries in the absence of knowledge returned by Retrieval Module. The results are shown in Table \ref{res_knowledge}. We can see that MIGRES with knowledge prompting achieves better results on Wikihop and NQ, showing that the LLM can indeed provide valuable information related to the input queries. The average iteration steps for MIGRES is only marginally reduced, this is because of the gpt's ability to refuse answering by responding "I don't know" in case of uncertainty or lack of knowledge.

We also evaluate the accuracy of the generated knowledge by checking if it contains the sub-answer with EM metric, the results are shown in Table \ref{acc_knowledge}. We can see that the generated knowledge gets a promising accuracy on Wikihop and Musique, thereby offering valuable information to address the input query. The low accuracy observed on the NQ cound stem from its answers being less standardized, resulting in reduced EM scores. For instance, gpt will generate \emph{French immigrants settled in various regions across Texas} for the question \emph{where did the french immigrants settle in texas}, while the ground truth answer is \emph{present - day southeastern Texas}.


\section{Related Works}


\noindent \textbf{Query optimisation in RAG} The optimization of user original queries is a critical area of focus \cite{gao2023retrieval} in RAG. Initial approaches have attempted to decompose multi-hop questions using rule-based methods and supervised models \cite{min2019multi, sun2020sparqa, Khot2021TextMN}, or expand the query itself through Generation Augmented Retrieval\cite{Shwartz2020UnsupervisedCQ, liu-etal-2022-generated}. However, these strategies often fail to pinpoint the gaps in the knowledge of language models

With the discovery of the reasoning capabilities inherent in LLMs \cite{Wei2022EmergentAO}, a series of studies represented by CoT \cite{Wei2022ChainOT} explored the use of the LLM to reform the query. These include static decomposition, where the original problem is dissected into sub-problems simultaneously \cite{zhou2022least, zhao2023verify}, but these methods lack flexibility. Therefore, more dynamic approaches have also been developed \cite{shao2023enhancing, feng2023retrievalgeneration, self-ask, yao2023react, kim2023tree}, which interact with external information sources in real-time. However, \citet{shao2023enhancing, feng2023retrievalgeneration} simply concatenate retrieved and generated content without clearly identifying the knowledge gaps in each iteration, \citet{self-ask, yao2023react} lacks the step of verification, and is easily misled by the retrieved useless information. \citet{kim2023tree} employ a tree structure for more detailed problem decomposition, which can be time-consuming. In contrast to the above methods, our approach prompts the language model to find the missing information and generates simple one-hop problems for more efficient retrieval, with less time costs.

\noindent \textbf{Retrieve-then-rerank framework} 
Retrieving documents that are relevant to the input query from the extensive pool of knowledge is inherently challenging \cite{gao2024retrievalaugmented, sun2024verifiable}, especially
when there exist irrelevant noise content throughout the context \cite{chen2024benchmarking, yoran2023making}. This noise not only wastes computational resources but also interferes with the generated content \cite{xu2024recomp}. Therefore, the retrieve-then-rerank paradigm is widely adopted to improve the quality of context by re-ranking retrieved knowledge to filter out the hard negative passages \cite{ma2023zero}. To streamline this process and condense the context, researchers suggest creating summaries or snippets pertinent \cite{gao-etal-2023-enabling, chen2023walking, xu2024recomp,  sun2024verifiable} to serve as knowledge augmentation. Nevertheless, \cite{gao-etal-2023-enabling, chen2023walking} do not assess the consistency between the retrieved text and the question, \cite{xu2024recomp, sun2024verifiable} are limited to coarse-grained reranking. Our approach enhances both aspects, by performing fine-grained filtering at the sentence level and by verifying the entailment between the top-ranked texts and the problem.

\section{Conclusion}
In this paper, we first experimentally verified the ability of LLMs to extract information as well as to know the missing information. Based on the discovery, we propose MIGRES, which leverages the missing information to steer new queries generation and subsequent knowledge retrieval, and thus facilitates the process of RAG for solving knowledge-intensive questions. Experimental results demonstrate the effectiveness of our propose method, which achieves superior or competitive performances compared with state-of-the-art baselines with generally less token consumption on external knowledge.


\clearpage
\section*{Limitation}
We also experimented with more advanced retrieval methods (e.g., utilizing bge \cite{bge_embedding} to conduct dense retrieval). Surprisingly, the performance of MIGRES with stronger retrieval was slighly inferior to that of BM25. This could be attributed to the ability of dense retrieval to recall knowledge that are semantically relevant but lack pertinent information. Such knowledge often covers similar topics or include the same nouns as the query. For instance, in response to the query \emph{When was the director of The House of Pulcini born?}, the dense retriever returns a passage titled \emph{Shari Springer Berman and Robert Pulcini} that talks about a team of filmmakers. The inclusion of such knowledge not only decreases the precision of information extraction \cite{cuconasu2024power}, but also leads to a less efficient iteration.


\bibliography{anthology,custom}

\begin{thebibliography}{44}
\expandafter\ifx\csname natexlab\endcsname\relax\def\natexlab#1{#1}\fi

\bibitem[{Bang et~al.(2023)Bang, Cahyawijaya, Lee, Dai, Su, Wilie, Lovenia, Ji, Yu, Chung, Do, Xu, and Fung}]{bang2023multitask}
Yejin Bang, Samuel Cahyawijaya, Nayeon Lee, Wenliang Dai, Dan Su, Bryan Wilie, Holy Lovenia, Ziwei Ji, Tiezheng Yu, Willy Chung, Quyet~V. Do, Yan Xu, and Pascale Fung. 2023.
\newblock \href {http://arxiv.org/abs/2302.04023} {A multitask, multilingual, multimodal evaluation of chatgpt on reasoning, hallucination, and interactivity}.

\bibitem[{Chen et~al.(2023)Chen, Pasunuru, Weston, and Celikyilmaz}]{chen2023walking}
Howard Chen, Ramakanth Pasunuru, Jason Weston, and Asli Celikyilmaz. 2023.
\newblock Walking down the memory maze: Beyond context limit through interactive reading.
\newblock \emph{arXiv preprint arXiv:2310.05029}.

\bibitem[{Chen et~al.(2024)Chen, Lin, Han, and Sun}]{chen2024benchmarking}
Jiawei Chen, Hongyu Lin, Xianpei Han, and Le~Sun. 2024.
\newblock Benchmarking large language models in retrieval-augmented generation.
\newblock In \emph{Proceedings of the AAAI Conference on Artificial Intelligence}, volume~38, pages 17754--17762.

\bibitem[{Cuconasu et~al.(2024)Cuconasu, Trappolini, Siciliano, Filice, Campagnano, Maarek, Tonellotto, and Silvestri}]{cuconasu2024power}
Florin Cuconasu, Giovanni Trappolini, Federico Siciliano, Simone Filice, Cesare Campagnano, Yoelle Maarek, Nicola Tonellotto, and Fabrizio Silvestri. 2024.
\newblock \href {http://arxiv.org/abs/2401.14887} {The power of noise: Redefining retrieval for rag systems}.

\bibitem[{Feng et~al.(2023)Feng, Feng, Zhao, Yang, and Qin}]{feng2023retrievalgeneration}
Zhangyin Feng, Xiaocheng Feng, Dezhi Zhao, Maojin Yang, and Bing Qin. 2023.
\newblock \href {http://arxiv.org/abs/2310.05149} {Retrieval-generation synergy augmented large language models}.

\bibitem[{Gao et~al.(2023{\natexlab{a}})Gao, Yen, Yu, and Chen}]{gao-etal-2023-enabling}
Tianyu Gao, Howard Yen, Jiatong Yu, and Danqi Chen. 2023{\natexlab{a}}.
\newblock \href {https://doi.org/10.18653/v1/2023.emnlp-main.398} {Enabling large language models to generate text with citations}.
\newblock In \emph{Proceedings of the 2023 Conference on Empirical Methods in Natural Language Processing}, pages 6465--6488, Singapore. Association for Computational Linguistics.

\bibitem[{Gao et~al.(2024)Gao, Xiong, Gao, Jia, Pan, Bi, Dai, Sun, Guo, Wang, and Wang}]{gao2024retrievalaugmented}
Yunfan Gao, Yun Xiong, Xinyu Gao, Kangxiang Jia, Jinliu Pan, Yuxi Bi, Yi~Dai, Jiawei Sun, Qianyu Guo, Meng Wang, and Haofen Wang. 2024.
\newblock \href {http://arxiv.org/abs/2312.10997} {Retrieval-augmented generation for large language models: A survey}.

\bibitem[{Gao et~al.(2023{\natexlab{b}})Gao, Xiong, Gao, Jia, Pan, Bi, Dai, Sun, and Wang}]{gao2023retrieval}
Yunfan Gao, Yun Xiong, Xinyu Gao, Kangxiang Jia, Jinliu Pan, Yuxi Bi, Yi~Dai, Jiawei Sun, and Haofen Wang. 2023{\natexlab{b}}.
\newblock Retrieval-augmented generation for large language models: A survey.
\newblock \emph{arXiv preprint arXiv:2312.10997}.

\bibitem[{Geva et~al.(2021)Geva, Khashabi, Segal, Khot, Roth, and Berant}]{geva2021strategyqa}
Mor Geva, Daniel Khashabi, Elad Segal, Tushar Khot, Dan Roth, and Jonathan Berant. 2021.
\newblock {Did Aristotle Use a Laptop? A Question Answering Benchmark with Implicit Reasoning Strategies}.
\newblock \emph{Transactions of the Association for Computational Linguistics (TACL)}.

\bibitem[{Ho et~al.(2020)Ho, Duong~Nguyen, Sugawara, and Aizawa}]{xanh2020_2wikimultihop}
Xanh Ho, Anh-Khoa Duong~Nguyen, Saku Sugawara, and Akiko Aizawa. 2020.
\newblock \href {https://www.aclweb.org/anthology/2020.coling-main.580} {Constructing a multi-hop {QA} dataset for comprehensive evaluation of reasoning steps}.
\newblock In \emph{Proceedings of the 28th International Conference on Computational Linguistics}, pages 6609--6625, Barcelona, Spain (Online). International Committee on Computational Linguistics.

\bibitem[{Honovich et~al.(2022)Honovich, Aharoni, Herzig, Taitelbaum, Kukliansy, Cohen, Scialom, Szpektor, Hassidim, and Matias}]{honovich-etal-2022-true-evaluating}
Or~Honovich, Roee Aharoni, Jonathan Herzig, Hagai Taitelbaum, Doron Kukliansy, Vered Cohen, Thomas Scialom, Idan Szpektor, Avinatan Hassidim, and Yossi Matias. 2022.
\newblock \href {https://doi.org/10.18653/v1/2022.naacl-main.287} {{TRUE}: Re-evaluating factual consistency evaluation}.
\newblock In \emph{Proceedings of the 2022 Conference of the North American Chapter of the Association for Computational Linguistics: Human Language Technologies}, pages 3905--3920, Seattle, United States. Association for Computational Linguistics.

\bibitem[{Huang et~al.(2023)Huang, Yu, Ma, Zhong, Feng, Wang, Chen, Peng, Feng, Qin, and Liu}]{huang2023survey}
Lei Huang, Weijiang Yu, Weitao Ma, Weihong Zhong, Zhangyin Feng, Haotian Wang, Qianglong Chen, Weihua Peng, Xiaocheng Feng, Bing Qin, and Ting Liu. 2023.
\newblock \href {http://arxiv.org/abs/2311.05232} {A survey on hallucination in large language models: Principles, taxonomy, challenges, and open questions}.

\bibitem[{Izacard et~al.(2022)Izacard, Lewis, Lomeli, Hosseini, Petroni, Schick, Dwivedi-Yu, Joulin, Riedel, and Grave}]{izacard2022atlas}
Gautier Izacard, Patrick Lewis, Maria Lomeli, Lucas Hosseini, Fabio Petroni, Timo Schick, Jane Dwivedi-Yu, Armand Joulin, Sebastian Riedel, and Edouard Grave. 2022.
\newblock \href {http://arxiv.org/abs/2208.03299} {Atlas: Few-shot learning with retrieval augmented language models}.

\bibitem[{Joshi et~al.(2017)Joshi, Choi, Weld, and Zettlemoyer}]{joshi-etal-2017-triviaqa}
Mandar Joshi, Eunsol Choi, Daniel Weld, and Luke Zettlemoyer. 2017.
\newblock \href {https://doi.org/10.18653/v1/P17-1147} {{T}rivia{QA}: A large scale distantly supervised challenge dataset for reading comprehension}.
\newblock In \emph{Proceedings of the 55th Annual Meeting of the Association for Computational Linguistics (Volume 1: Long Papers)}, pages 1601--1611, Vancouver, Canada. Association for Computational Linguistics.

\bibitem[{Karpukhin et~al.(2020)Karpukhin, Oğuz, Min, Lewis, Wu, Edunov, Chen, and tau Yih}]{karpukhin2020dense}
Vladimir Karpukhin, Barlas Oğuz, Sewon Min, Patrick Lewis, Ledell Wu, Sergey Edunov, Danqi Chen, and Wen tau Yih. 2020.
\newblock \href {http://arxiv.org/abs/2004.04906} {Dense passage retrieval for open-domain question answering}.

\bibitem[{Khandelwal et~al.(2020)Khandelwal, Levy, Jurafsky, Zettlemoyer, and Lewis}]{Khandelwal2020Generalization}
Urvashi Khandelwal, Omer Levy, Dan Jurafsky, Luke Zettlemoyer, and Mike Lewis. 2020.
\newblock \href {https://openreview.net/forum?id=HklBjCEKvH} {Generalization through memorization: Nearest neighbor language models}.
\newblock In \emph{International Conference on Learning Representations}.

\bibitem[{Khot et~al.(2021)Khot, Khashabi, Richardson, Clark, and Sabharwal}]{Khot2021TextMN}
Tushar Khot, Daniel Khashabi, Kyle Richardson, Peter Clark, and Ashish Sabharwal. 2021.
\newblock \href {https://api.semanticscholar.org/CorpusID:221448158} {Text modular networks: Learning to decompose tasks in the language of existing models}.
\newblock \emph{Proceedings of the 2021 Conference of the North American Chapter of the Association for Computational Linguistics: Human Language Technologies}, page 1264–1279.

\bibitem[{Kim et~al.(2023)Kim, Kim, Jeon, Park, and Kang}]{kim2023tree}
Gangwoo Kim, Sungdong Kim, Byeongguk Jeon, Joonsuk Park, and Jaewoo Kang. 2023.
\newblock Tree of clarifications: Answering ambiguous questions with retrieval-augmented large language models.
\newblock In \emph{Proceedings of the 2023 Conference on Empirical Methods in Natural Language Processing}, pages 996--1009.

\bibitem[{Kwiatkowski et~al.(2019)Kwiatkowski, Palomaki, Redfield, Collins, Parikh, Alberti, Epstein, Polosukhin, Devlin, Lee, Toutanova, Jones, Kelcey, Chang, Dai, Uszkoreit, Le, and Petrov}]{kwiatkowski-etal-2019-natural}
Tom Kwiatkowski, Jennimaria Palomaki, Olivia Redfield, Michael Collins, Ankur Parikh, Chris Alberti, Danielle Epstein, Illia Polosukhin, Jacob Devlin, Kenton Lee, Kristina Toutanova, Llion Jones, Matthew Kelcey, Ming-Wei Chang, Andrew~M. Dai, Jakob Uszkoreit, Quoc Le, and Slav Petrov. 2019.
\newblock \href {https://doi.org/10.1162/tacl_a_00276} {Natural questions: A benchmark for question answering research}.
\newblock \emph{Transactions of the Association for Computational Linguistics}, 7:452--466.

\bibitem[{Liu et~al.(2022)Liu, Liu, Lu, Welleck, West, Le~Bras, Choi, and Hajishirzi}]{liu-etal-2022-generated}
Jiacheng Liu, Alisa Liu, Ximing Lu, Sean Welleck, Peter West, Ronan Le~Bras, Yejin Choi, and Hannaneh Hajishirzi. 2022.
\newblock \href {https://doi.org/10.18653/v1/2022.acl-long.225} {Generated knowledge prompting for commonsense reasoning}.
\newblock In \emph{Proceedings of the 60th Annual Meeting of the Association for Computational Linguistics (Volume 1: Long Papers)}, pages 3154--3169, Dublin, Ireland. Association for Computational Linguistics.

\bibitem[{Ma et~al.(2023)Ma, Zhang, Pradeep, and Lin}]{ma2023zero}
Xueguang Ma, Xinyu Zhang, Ronak Pradeep, and Jimmy Lin. 2023.
\newblock Zero-shot listwise document reranking with a large language model.
\newblock \emph{arXiv preprint arXiv:2305.02156}.

\bibitem[{Min et~al.(2019)Min, Zhong, Zettlemoyer, and Hajishirzi}]{min2019multi}
Sewon Min, Victor Zhong, Luke Zettlemoyer, and Hannaneh Hajishirzi. 2019.
\newblock Multi-hop reading comprehension through question decomposition and rescoring.
\newblock \emph{arXiv preprint arXiv:1906.02916}.

\bibitem[{OpenAI et~al.(2023)OpenAI, :, Achiam, Adler, Agarwal, Ahmad, Akkaya, Aleman, Almeida, Altenschmidt, Altman, Anadkat, Avila, Babuschkin, Balaji, Balcom, Baltescu, Bao, Bavarian, Belgum, Bello, Berdine, Bernadett-Shapiro, Berner, Bogdonoff, Boiko, Boyd, Brakman, Brockman, Brooks, Brundage, Button, Cai, Campbell, Cann, Carey, Carlson, Carmichael, Chan, Chang, Chantzis, Chen, Chen, Chen, Chen, Chen, Chess, Cho, Chu, Chung, Cummings, Currier, Dai, Decareaux, Degry, Deutsch, Deville, Dhar, Dohan, Dowling, Dunning, Ecoffet, Eleti, Eloundou, Farhi, Fedus, Felix, Fishman, Forte, Fulford, Gao, Georges, Gibson, Goel, Gogineni, Goh, Gontijo-Lopes, Gordon, Grafstein, Gray, Greene, Gross, Gu, Guo, Hallacy, Han, Harris, He, Heaton, Heidecke, Hesse, Hickey, Hickey, Hoeschele, Houghton, Hsu, Hu, Hu, Huizinga, Jain, Jain, Jang, Jiang, Jiang, Jin, Jin, Jomoto, Jonn, Jun, Kaftan, Łukasz Kaiser, Kamali, Kanitscheider, Keskar, Khan, Kilpatrick, Kim, Kim, Kim, Kirchner, Kiros, Knight, Kokotajlo, Łukasz Kondraciuk,
  Kondrich, Konstantinidis, Kosic, Krueger, Kuo, Lampe, Lan, Lee, Leike, Leung, Levy, Li, Lim, Lin, Lin, Litwin, Lopez, Lowe, Lue, Makanju, Malfacini, Manning, Markov, Markovski, Martin, Mayer, Mayne, McGrew, McKinney, McLeavey, McMillan, McNeil, Medina, Mehta, Menick, Metz, Mishchenko, Mishkin, Monaco, Morikawa, Mossing, Mu, Murati, Murk, Mély, Nair, Nakano, Nayak, Neelakantan, Ngo, Noh, Ouyang, O'Keefe, Pachocki, Paino, Palermo, Pantuliano, Parascandolo, Parish, Parparita, Passos, Pavlov, Peng, Perelman, de~Avila Belbute~Peres, Petrov, de~Oliveira~Pinto, Michael, Pokorny, Pokrass, Pong, Powell, Power, Power, Proehl, Puri, Radford, Rae, Ramesh, Raymond, Real, Rimbach, Ross, Rotsted, Roussez, Ryder, Saltarelli, Sanders, Santurkar, Sastry, Schmidt, Schnurr, Schulman, Selsam, Sheppard, Sherbakov, Shieh, Shoker, Shyam, Sidor, Sigler, Simens, Sitkin, Slama, Sohl, Sokolowsky, Song, Staudacher, Such, Summers, Sutskever, Tang, Tezak, Thompson, Tillet, Tootoonchian, Tseng, Tuggle, Turley, Tworek, Uribe, Vallone,
  Vijayvergiya, Voss, Wainwright, Wang, Wang, Wang, Ward, Wei, Weinmann, Welihinda, Welinder, Weng, Weng, Wiethoff, Willner, Winter, Wolrich, Wong, Workman, Wu, Wu, Wu, Xiao, Xu, Yoo, Yu, Yuan, Zaremba, Zellers, Zhang, Zhang, Zhao, Zheng, Zhuang, Zhuk, and Zoph}]{openai2023gpt4}
OpenAI, :, Josh Achiam, Steven Adler, Sandhini Agarwal, Lama Ahmad, Ilge Akkaya, Florencia~Leoni Aleman, Diogo Almeida, Janko Altenschmidt, Sam Altman, Shyamal Anadkat, Red Avila, Igor Babuschkin, Suchir Balaji, Valerie Balcom, Paul Baltescu, Haiming Bao, Mo~Bavarian, Jeff Belgum, Irwan Bello, Jake Berdine, Gabriel Bernadett-Shapiro, Christopher Berner, Lenny Bogdonoff, Oleg Boiko, Madelaine Boyd, Anna-Luisa Brakman, Greg Brockman, Tim Brooks, Miles Brundage, Kevin Button, Trevor Cai, Rosie Campbell, Andrew Cann, Brittany Carey, Chelsea Carlson, Rory Carmichael, Brooke Chan, Che Chang, Fotis Chantzis, Derek Chen, Sully Chen, Ruby Chen, Jason Chen, Mark Chen, Ben Chess, Chester Cho, Casey Chu, Hyung~Won Chung, Dave Cummings, Jeremiah Currier, Yunxing Dai, Cory Decareaux, Thomas Degry, Noah Deutsch, Damien Deville, Arka Dhar, David Dohan, Steve Dowling, Sheila Dunning, Adrien Ecoffet, Atty Eleti, Tyna Eloundou, David Farhi, Liam Fedus, Niko Felix, Simón~Posada Fishman, Juston Forte, Isabella Fulford, Leo Gao,
  Elie Georges, Christian Gibson, Vik Goel, Tarun Gogineni, Gabriel Goh, Rapha Gontijo-Lopes, Jonathan Gordon, Morgan Grafstein, Scott Gray, Ryan Greene, Joshua Gross, Shixiang~Shane Gu, Yufei Guo, Chris Hallacy, Jesse Han, Jeff Harris, Yuchen He, Mike Heaton, Johannes Heidecke, Chris Hesse, Alan Hickey, Wade Hickey, Peter Hoeschele, Brandon Houghton, Kenny Hsu, Shengli Hu, Xin Hu, Joost Huizinga, Shantanu Jain, Shawn Jain, Joanne Jang, Angela Jiang, Roger Jiang, Haozhun Jin, Denny Jin, Shino Jomoto, Billie Jonn, Heewoo Jun, Tomer Kaftan, Łukasz Kaiser, Ali Kamali, Ingmar Kanitscheider, Nitish~Shirish Keskar, Tabarak Khan, Logan Kilpatrick, Jong~Wook Kim, Christina Kim, Yongjik Kim, Hendrik Kirchner, Jamie Kiros, Matt Knight, Daniel Kokotajlo, Łukasz Kondraciuk, Andrew Kondrich, Aris Konstantinidis, Kyle Kosic, Gretchen Krueger, Vishal Kuo, Michael Lampe, Ikai Lan, Teddy Lee, Jan Leike, Jade Leung, Daniel Levy, Chak~Ming Li, Rachel Lim, Molly Lin, Stephanie Lin, Mateusz Litwin, Theresa Lopez, Ryan Lowe,
  Patricia Lue, Anna Makanju, Kim Malfacini, Sam Manning, Todor Markov, Yaniv Markovski, Bianca Martin, Katie Mayer, Andrew Mayne, Bob McGrew, Scott~Mayer McKinney, Christine McLeavey, Paul McMillan, Jake McNeil, David Medina, Aalok Mehta, Jacob Menick, Luke Metz, Andrey Mishchenko, Pamela Mishkin, Vinnie Monaco, Evan Morikawa, Daniel Mossing, Tong Mu, Mira Murati, Oleg Murk, David Mély, Ashvin Nair, Reiichiro Nakano, Rajeev Nayak, Arvind Neelakantan, Richard Ngo, Hyeonwoo Noh, Long Ouyang, Cullen O'Keefe, Jakub Pachocki, Alex Paino, Joe Palermo, Ashley Pantuliano, Giambattista Parascandolo, Joel Parish, Emy Parparita, Alex Passos, Mikhail Pavlov, Andrew Peng, Adam Perelman, Filipe de~Avila Belbute~Peres, Michael Petrov, Henrique~Ponde de~Oliveira~Pinto, Michael, Pokorny, Michelle Pokrass, Vitchyr Pong, Tolly Powell, Alethea Power, Boris Power, Elizabeth Proehl, Raul Puri, Alec Radford, Jack Rae, Aditya Ramesh, Cameron Raymond, Francis Real, Kendra Rimbach, Carl Ross, Bob Rotsted, Henri Roussez, Nick Ryder,
  Mario Saltarelli, Ted Sanders, Shibani Santurkar, Girish Sastry, Heather Schmidt, David Schnurr, John Schulman, Daniel Selsam, Kyla Sheppard, Toki Sherbakov, Jessica Shieh, Sarah Shoker, Pranav Shyam, Szymon Sidor, Eric Sigler, Maddie Simens, Jordan Sitkin, Katarina Slama, Ian Sohl, Benjamin Sokolowsky, Yang Song, Natalie Staudacher, Felipe~Petroski Such, Natalie Summers, Ilya Sutskever, Jie Tang, Nikolas Tezak, Madeleine Thompson, Phil Tillet, Amin Tootoonchian, Elizabeth Tseng, Preston Tuggle, Nick Turley, Jerry Tworek, Juan Felipe~Cerón Uribe, Andrea Vallone, Arun Vijayvergiya, Chelsea Voss, Carroll Wainwright, Justin~Jay Wang, Alvin Wang, Ben Wang, Jonathan Ward, Jason Wei, CJ~Weinmann, Akila Welihinda, Peter Welinder, Jiayi Weng, Lilian Weng, Matt Wiethoff, Dave Willner, Clemens Winter, Samuel Wolrich, Hannah Wong, Lauren Workman, Sherwin Wu, Jeff Wu, Michael Wu, Kai Xiao, Tao Xu, Sarah Yoo, Kevin Yu, Qiming Yuan, Wojciech Zaremba, Rowan Zellers, Chong Zhang, Marvin Zhang, Shengjia Zhao, Tianhao
  Zheng, Juntang Zhuang, William Zhuk, and Barret Zoph. 2023.
\newblock \href {http://arxiv.org/abs/2303.08774} {Gpt-4 technical report}.

\bibitem[{Ouyang et~al.(2022)Ouyang, Wu, Jiang, Almeida, Wainwright, Mishkin, Zhang, Agarwal, Slama, Ray, Schulman, Hilton, Kelton, Miller, Simens, Askell, Welinder, Christiano, Leike, and Lowe}]{ouyang2022training}
Long Ouyang, Jeff Wu, Xu~Jiang, Diogo Almeida, Carroll~L. Wainwright, Pamela Mishkin, Chong Zhang, Sandhini Agarwal, Katarina Slama, Alex Ray, John Schulman, Jacob Hilton, Fraser Kelton, Luke Miller, Maddie Simens, Amanda Askell, Peter Welinder, Paul Christiano, Jan Leike, and Ryan Lowe. 2022.
\newblock \href {http://arxiv.org/abs/2203.02155} {Training language models to follow instructions with human feedback}.

\bibitem[{Press et~al.(2023)Press, Zhang, Min, Schmidt, Smith, and Lewis}]{self-ask}
Ofir Press, Muru Zhang, Sewon Min, Ludwig Schmidt, Noah~A. Smith, and Mike Lewis. 2023.
\newblock \href {http://arxiv.org/abs/2210.03350} {Measuring and narrowing the compositionality gap in language models}.

\bibitem[{Shao et~al.(2023{\natexlab{a}})Shao, Gong, Shen, Huang, Duan, and Chen}]{shao-etal-2023-enhancing}
Zhihong Shao, Yeyun Gong, Yelong Shen, Minlie Huang, Nan Duan, and Weizhu Chen. 2023{\natexlab{a}}.
\newblock \href {https://doi.org/10.18653/v1/2023.findings-emnlp.620} {Enhancing retrieval-augmented large language models with iterative retrieval-generation synergy}.
\newblock In \emph{Findings of the Association for Computational Linguistics: EMNLP 2023}, pages 9248--9274, Singapore. Association for Computational Linguistics.

\bibitem[{Shao et~al.(2023{\natexlab{b}})Shao, Gong, Shen, Huang, Duan, and Chen}]{shao2023enhancing}
Zhihong Shao, Yeyun Gong, Yelong Shen, Minlie Huang, Nan Duan, and Weizhu Chen. 2023{\natexlab{b}}.
\newblock Enhancing retrieval-augmented large language models with iterative retrieval-generation synergy.
\newblock In \emph{Findings of the Association for Computational Linguistics: EMNLP 2023}, pages 9248--9274.

\bibitem[{Shwartz et~al.(2020)Shwartz, West, Bras, Bhagavatula, and Choi}]{Shwartz2020UnsupervisedCQ}
Vered Shwartz, Peter West, Ronan~Le Bras, Chandra Bhagavatula, and Yejin Choi. 2020.
\newblock \href {https://api.semanticscholar.org/CorpusID:215745286} {Unsupervised commonsense question answering with self-talk}.
\newblock In \emph{Conference on Empirical Methods in Natural Language Processing}.

\bibitem[{Sun et~al.(2024)Sun, Cai, Wang, Hou, Wei, Wang, Zhang, and Yin}]{sun2024verifiable}
Hao Sun, Hengyi Cai, Bo~Wang, Yingyan Hou, Xiaochi Wei, Shuaiqiang Wang, Yan Zhang, and Dawei Yin. 2024.
\newblock \href {http://arxiv.org/abs/2312.09075} {Towards verifiable text generation with evolving memory and self-reflection}.

\bibitem[{Sun et~al.(2020)Sun, Zhang, Cheng, and Qu}]{sun2020sparqa}
Yawei Sun, Lingling Zhang, Gong Cheng, and Yuzhong Qu. 2020.
\newblock Sparqa: skeleton-based semantic parsing for complex questions over knowledge bases.
\newblock In \emph{Proceedings of the AAAI conference on artificial intelligence}, volume~34, pages 8952--8959.

\bibitem[{Touvron et~al.(2023)Touvron, Martin, Stone, Albert, Almahairi, Babaei, Bashlykov, Batra, Bhargava, Bhosale, Bikel, Blecher, Ferrer, Chen, Cucurull, Esiobu, Fernandes, Fu, Fu, Fuller, Gao, Goswami, Goyal, Hartshorn, Hosseini, Hou, Inan, Kardas, Kerkez, Khabsa, Kloumann, Korenev, Koura, Lachaux, Lavril, Lee, Liskovich, Lu, Mao, Martinet, Mihaylov, Mishra, Molybog, Nie, Poulton, Reizenstein, Rungta, Saladi, Schelten, Silva, Smith, Subramanian, Tan, Tang, Taylor, Williams, Kuan, Xu, Yan, Zarov, Zhang, Fan, Kambadur, Narang, Rodriguez, Stojnic, Edunov, and Scialom}]{touvron2023llama}
Hugo Touvron, Louis Martin, Kevin Stone, Peter Albert, Amjad Almahairi, Yasmine Babaei, Nikolay Bashlykov, Soumya Batra, Prajjwal Bhargava, Shruti Bhosale, Dan Bikel, Lukas Blecher, Cristian~Canton Ferrer, Moya Chen, Guillem Cucurull, David Esiobu, Jude Fernandes, Jeremy Fu, Wenyin Fu, Brian Fuller, Cynthia Gao, Vedanuj Goswami, Naman Goyal, Anthony Hartshorn, Saghar Hosseini, Rui Hou, Hakan Inan, Marcin Kardas, Viktor Kerkez, Madian Khabsa, Isabel Kloumann, Artem Korenev, Punit~Singh Koura, Marie-Anne Lachaux, Thibaut Lavril, Jenya Lee, Diana Liskovich, Yinghai Lu, Yuning Mao, Xavier Martinet, Todor Mihaylov, Pushkar Mishra, Igor Molybog, Yixin Nie, Andrew Poulton, Jeremy Reizenstein, Rashi Rungta, Kalyan Saladi, Alan Schelten, Ruan Silva, Eric~Michael Smith, Ranjan Subramanian, Xiaoqing~Ellen Tan, Binh Tang, Ross Taylor, Adina Williams, Jian~Xiang Kuan, Puxin Xu, Zheng Yan, Iliyan Zarov, Yuchen Zhang, Angela Fan, Melanie Kambadur, Sharan Narang, Aurelien Rodriguez, Robert Stojnic, Sergey Edunov, and Thomas
  Scialom. 2023.
\newblock \href {http://arxiv.org/abs/2307.09288} {Llama 2: Open foundation and fine-tuned chat models}.

\bibitem[{Trivedi et~al.(2022)Trivedi, Balasubramanian, Khot, and Sabharwal}]{trivedi2021musique}
Harsh Trivedi, Niranjan Balasubramanian, Tushar Khot, and Ashish Sabharwal. 2022.
\newblock {M}u{S}i{Q}ue: Multihop questions via single-hop question composition.
\newblock \emph{Transactions of the Association for Computational Linguistics}.

\bibitem[{Wang et~al.(2023)Wang, Duan, Wang, Li, Xian, Yin, Rong, and Xiong}]{wang2023knowledgedriven}
Keheng Wang, Feiyu Duan, Sirui Wang, Peiguang Li, Yunsen Xian, Chuantao Yin, Wenge Rong, and Zhang Xiong. 2023.
\newblock \href {http://arxiv.org/abs/2308.13259} {Knowledge-driven cot: Exploring faithful reasoning in llms for knowledge-intensive question answering}.

\bibitem[{Wei et~al.(2022{\natexlab{a}})Wei, Tay, Bommasani, Raffel, Zoph, Borgeaud, Yogatama, Bosma, Zhou, Metzler, hsin Chi, Hashimoto, Vinyals, Liang, Dean, and Fedus}]{Wei2022EmergentAO}
Jason Wei, Yi~Tay, Rishi Bommasani, Colin Raffel, Barret Zoph, Sebastian Borgeaud, Dani Yogatama, Maarten Bosma, Denny Zhou, Donald Metzler, Ed~Huai hsin Chi, Tatsunori Hashimoto, Oriol Vinyals, Percy Liang, Jeff Dean, and William Fedus. 2022{\natexlab{a}}.
\newblock \href {https://api.semanticscholar.org/CorpusID:249674500} {Emergent abilities of large language models}.
\newblock \emph{ArXiv}, abs/2206.07682.

\bibitem[{Wei et~al.(2022{\natexlab{b}})Wei, Wang, Schuurmans, Bosma, hsin Chi, Xia, Le, and Zhou}]{Wei2022ChainOT}
Jason Wei, Xuezhi Wang, Dale Schuurmans, Maarten Bosma, Ed~Huai hsin Chi, F.~Xia, Quoc Le, and Denny Zhou. 2022{\natexlab{b}}.
\newblock \href {https://api.semanticscholar.org/CorpusID:246411621} {Chain of thought prompting elicits reasoning in large language models}.
\newblock \emph{ArXiv}, abs/2201.11903.

\bibitem[{Wei et~al.(2022{\natexlab{c}})Wei, Wang, Schuurmans, Bosma, Xia, Chi, Le, Zhou et~al.}]{wei2022chain}
Jason Wei, Xuezhi Wang, Dale Schuurmans, Maarten Bosma, Fei Xia, Ed~Chi, Quoc~V Le, Denny Zhou, et~al. 2022{\natexlab{c}}.
\newblock Chain-of-thought prompting elicits reasoning in large language models.
\newblock \emph{Advances in neural information processing systems}, 35:24824--24837.

\bibitem[{Xiao et~al.(2023)Xiao, Liu, Zhang, and Muennighoff}]{bge_embedding}
Shitao Xiao, Zheng Liu, Peitian Zhang, and Niklas Muennighoff. 2023.
\newblock \href {http://arxiv.org/abs/2309.07597} {C-pack: Packaged resources to advance general chinese embedding}.

\bibitem[{Xu et~al.(2024)Xu, Shi, and Choi}]{xu2024recomp}
Fangyuan Xu, Weijia Shi, and Eunsol Choi. 2024.
\newblock \href {https://openreview.net/forum?id=mlJLVigNHp} {{RECOMP}: Improving retrieval-augmented {LM}s with context compression and selective augmentation}.
\newblock In \emph{The Twelfth International Conference on Learning Representations}.

\bibitem[{Yang et~al.(2018)Yang, Qi, Zhang, Bengio, Cohen, Salakhutdinov, and Manning}]{yang-etal-2018-hotpotqa}
Zhilin Yang, Peng Qi, Saizheng Zhang, Yoshua Bengio, William Cohen, Ruslan Salakhutdinov, and Christopher~D. Manning. 2018.
\newblock \href {https://doi.org/10.18653/v1/D18-1259} {{H}otpot{QA}: A dataset for diverse, explainable multi-hop question answering}.
\newblock In \emph{Proceedings of the 2018 Conference on Empirical Methods in Natural Language Processing}, pages 2369--2380, Brussels, Belgium. Association for Computational Linguistics.

\bibitem[{Yao et~al.(2023)Yao, Zhao, Yu, Du, Shafran, Narasimhan, and Cao}]{yao2023react}
Shunyu Yao, Jeffrey Zhao, Dian Yu, Nan Du, Izhak Shafran, Karthik~R Narasimhan, and Yuan Cao. 2023.
\newblock \href {https://openreview.net/forum?id=WE_vluYUL-X} {React: Synergizing reasoning and acting in language models}.
\newblock In \emph{The Eleventh International Conference on Learning Representations}.

\bibitem[{Yoran et~al.(2023{\natexlab{a}})Yoran, Wolfson, Bogin, Katz, Deutch, and Berant}]{yoran2023answering}
Ori Yoran, Tomer Wolfson, Ben Bogin, Uri Katz, Daniel Deutch, and Jonathan Berant. 2023{\natexlab{a}}.
\newblock \href {http://arxiv.org/abs/2304.13007} {Answering questions by meta-reasoning over multiple chains of thought}.

\bibitem[{Yoran et~al.(2023{\natexlab{b}})Yoran, Wolfson, Ram, and Berant}]{yoran2023making}
Ori Yoran, Tomer Wolfson, Ori Ram, and Jonathan Berant. 2023{\natexlab{b}}.
\newblock \href {http://arxiv.org/abs/2310.01558} {Making retrieval-augmented language models robust to irrelevant context}.

\bibitem[{Zhao et~al.(2023)Zhao, Li, Joty, Qin, and Bing}]{zhao2023verify}
Ruochen Zhao, Xingxuan Li, Shafiq Joty, Chengwei Qin, and Lidong Bing. 2023.
\newblock Verify-and-edit: A knowledge-enhanced chain-of-thought framework.
\newblock In \emph{Proceedings of the 61st Annual Meeting of the Association for Computational Linguistics (Volume 1: Long Papers)}, pages 5823--5840.

\bibitem[{Zhou et~al.(2022)Zhou, Sch{\"a}rli, Hou, Wei, Scales, Wang, Schuurmans, Cui, Bousquet, Le et~al.}]{zhou2022least}
Denny Zhou, Nathanael Sch{\"a}rli, Le~Hou, Jason Wei, Nathan Scales, Xuezhi Wang, Dale Schuurmans, Claire Cui, Olivier Bousquet, Quoc~V Le, et~al. 2022.
\newblock Least-to-most prompting enables complex reasoning in large language models.
\newblock In \emph{The Eleventh International Conference on Learning Representations}.

\end{thebibliography}
\bibliographystyle{acl_natbib}

\clearpage
\onecolumn
\appendix

\section{Prompt for QA pairs Decomposing}

\begin{longtable}[ht]{p{16cm}}
    \toprule
    \textbf{\large \# Useful information extraction} \\
    \\
    Given the following question with its final answer and evidence, please generate the sub-questions corresponding to the evidence to complete the question decomposition. \\
    \\
    \textbf{\large \# Wikihop} \\
    Question: Which film was released more recently, Fatal Lady or Every Blessed Day? \\
    Final answer: Every Blessed Day \\ 
    Evidence: {[['Fatal Lady', 'publication date', '1936'], ['Every BlessedDay', 'publication date', '2012']]} \\
    \\
    Sub-question 1: When was Fatal Lady released? \\
    Sub-answer 1: Fatal Lady was released in 1936 \\
    Sub-question 2: When was Every Blessed Day released? \\
    Sub-answer 2: Every Blessed Day was released in 2012 \\
    \\
    \textbf{\large \# Musique} \\
    Question: How long is a governor's term in the state where Selma, Lord, Selma takes place? \\
    Final answer: Four years \\
    Evidence: {[\{"question": 'Which place is Selma, Lord, Selma in?', "answer": 'Alabama'\}, \{"question": "how long is a governor's term in \#1", "answer: 'Four years'\}]} \\
    Sub-question 1: Which place is Selma, Lord, Selma in? \\
    Sub-answer 1: The place Selma, Lord, Selma is located in is Alabama. \\
    Sub-question 2: how long is a governor's term in Alabama \\
    Sub-answer 2: The governor's term in Alabama is four years. \\
    \bottomrule
    \caption{Prompts and cases for QA pairs decomposing.}
    \label{prompt_decom}
\end{longtable}

\section{Prompts and cases in preliminary experiment}
\label{appendix_prompt_prelimary}

\begin{longtable}[ht]{p{16cm}}
    \toprule
    \textbf{\large \# Useful information extraction} \\
    \\
Given the following question and passages, please distillate useful information from
the Passages to address the Question effectively and list the support passage index for each distilled information. Your response should be under the format \{"useful\_information": [\{"info": statement of useful information, "support\_passages": [indexes of support passages]\}]\}. The provided passages might be irrelevant and contain no useful information. Not provided information should not appear in your response. Please generate a dict format response. \\
\\
Question: Who is the director of Golmaal (2008 Film)? \\
\\
Passage 0: (Title: Les Oreilles) Les Oreilles is a 2008 film. \\
Passage 1: (Title: Henry Moore (cricketer)) Henry Walter Moore( 1849 – 20 August 1916) was an English- born first-class cricketer who spent most of his life in New Zealand. \\
Passage 2: (Title: Arugba) Arugba is a 2008 film. \\
Passage 3: (Title: Swapan Saha) Swapan Saha( born 10 January 1930 in Ajmer, Rajasthan, India) is an Indian Bengali film director, producer, story writer and score composer. \\
Passage 4: (Title: Terence Macartney-Filgate) Terence Macartney-Filgate (born August 6, 1924 in England, United Kingdom) is a British-Canadian film director who has directed, written, produced or shot more than 100 films in a career spanning more than 50 years. \\
\\
Your response:

\{"useful\_information": [\{"info": "Swapan Saha is the director of Golmaal (2008 Film)", "support\_passages": [3]\}]\} \\
\\
\# \textbf{\large{Verification of usefulness}} \\
\\
Determine whether the provided information is useful for answering the question. Please answer "yes" or "no". \\
Information: Swapan Saha is the director of Golmaal \\
Question: Who is the director of Golmaal (2008 Film)? \\
\\
Yes. \\
    \bottomrule
    \caption{Prompts and cases for useful information extraction.}
    \label{useful prompt}
\end{longtable}

\begin{longtable}[ht]{p{16cm}}
    \toprule
    \textbf{\large \# Missing information generation} \\
    \\
Answer the question based on the provided information. If the question can not be answered, the answer should be "unanswerable" and you should give a summary of the missing information. Otherwise, provide your answer to the question. Your response should be under the format \{"answer": your answer, "missing\_information": the summary of the missing information\}. \\
\\
Question: What latitude is defined as being the border of the continent where Smokinya Cove is located? \\
\\
Information: Barcelona beat Manchester United to win the 2008-09 Champions League title. \\
\\
Your response: \{"answer": "unanswerable", "missing\_information": "The latitude of the border where Smokinya Cove is located."\} \\
\\
\# \textbf{\large{Verification of missing information}} \\
\\
Given the missing information and the candidate questions. List the index of questions aligned with the missing information, then provide your explanation. \\
\\
Missing information: The latitude of the border where Smokinya Cove is located. \\
\\
Candidate questions: \\
(\# Sub-questions pre-generated, as illustrated in Figure \ref{prompt_decom}.) \\
\\
1. What latitude is defined as being Antarctica 's border?\\
\\
Index: [1] \\
Explanation: The missing information relates to the latitude of the border where Smokinya Cove is located. The candidate question 1 directly asks about the latitude defining Antarctica's border, which aligns with the missing information. Therefore, the index for the aligned question is [1].\\
    \bottomrule
    \caption{Prompts and cases for missing information generation.}
    \label{missing prompt}
\end{longtable}

\section{Prompt for GPT knowledge prompting and answer evaluation}
\begin{longtable}[ht]{p{16cm}}
    \toprule
    \textbf{\large \# Prompt for knowledge prompting} \\
    Generation relevant information to the given question. \\
    Question: \{QUESTION\} \\
    Information: \\
    \\
    \textbf{\large \# Prompt for Evaluating the Correctness of a
Model Output} \\
    In the following task, you are given a Question, a model Prediction for the Question, and a Ground-truth Answer to the Question. You should decide whether the model Prediction implies the Ground-truth Answer. \\
    \\
    Question \\
    \{question\} \\
    Prediction \\
    \{model output\} \\
    Ground-truth Answer \\
    \{answer\} \\
    Does the Prediction imply the Ground-truth Answer? Output Yes or No: \\
    \bottomrule
    \caption{Prompts for GPT knowledge prompting and for evaluating the correctness of a model output.}
    \label{other_prompt}
\end{longtable}

\section{Hyper-parameter settings for MIGRES}
We list the hyper-parameter settings of MIGRES in Table \ref{hyper-setting}.
\begin{table*}[h]
    \centering
    \scalebox{1.0}{
    \begin{tabular}{c|ccc}
    \toprule
    \textbf{Dataset} & \textbf{Multi-hop QA} & \textbf{ODQA} & \textbf{Commonsense QA}\\
    \hline
    Max Iteration Steps $\mathcal{T}$ & 5 & 3 & 5 \\
    Relevance Threshold $\delta$ & 3.0 & 5.0 & 0.0 \\
    Max Num of Passages Concatenated $k$ & 5 & 5 & 5 \\
    Nums of Passages Retrieved for each Query & 50 & 50 & 50 \\
    
    \bottomrule
    \end{tabular}}
    \caption{Hyper-parameter settings for MIGRES. We predetermine the relevance threshold $\delta$ utilizing the training set, where we conduct knowledge retrieval with the original questions on a sub-sampled set, and compute relevance scores of the retrieved passages. We adjust $\delta$ so that more than 80\% of the questions find at least one external passage with a relevance higher than this threshold.}
    \label{hyper-setting}
\end{table*}

\section{Comparison of different knowledge compression method}
\begin{longtable}[ht]{p{16cm}}
    \toprule
    \textbf{\large \# Comparison with different compression method} \\
    \\
<Question>
Which film has the director who died later, What Will You Do When You Catch Me? or School On Fire? \\
\\
<Retrieved passage>
(Title: School on Fire) School on Fire School on Fire is a 1988 Hong Kong action film directed by Ringo Lam. The film involves a young schoolgirl Chu Yuen Fong (Fennie Yuen) who becomes caught in a tragic stranglehold of triad activity after she testifies over a triad beating. When this news reaches the triad leader Brother Smart (Roy Cheung), Yuen Fong must pay him protection money for what she has done as events begin to escalate. The film involves a young schoolgirl Chu Yuen Fong (Fennie Yuen) who becomes caught in a tragic stranglehold of triad activity after she testifies over a \\
\\
\textbf{ \# SUMM} \\
School on Fire" is a 1988 Hong Kong action film directed by Ringo Lam. It follows Chu Yuen Fong, who gets involved in triad activity after testifying against them. The director, Ringo Lam, died later. It's irrelevant to the question about "What Will You Do When You Catch Me? \\
\\
\textbf{ \# SNIPPET} \\
Irrelevant. \\
\\
\textbf{ \# Sentence-level Filtering} \\
School on Fire School on Fire is a 1988 Hong Kong action film directed by Ringo Lam.\\
    \bottomrule
    \caption{Prompts and cases for missing information generation.}
    \label{compress_res}
\end{longtable}

\section{Prompts and cases for MIGRES}

\begin{longtable}[ht]{p{16cm}}
    \toprule
    \textbf{\large Cases of MIGRES} \\
    \\
\textbf{\large \# Case 1} \\
<Original Question> \\
Do both films Levity (Film) and I Come With The Rain have the directors that share the same nationality? \\
<Labels> \\
{\{"answer": "No", "evidences": [["Levity (film)", "director", "Ed Solomon"], ["I Come with the Rain", "director", "Tran Anh Hung"], ["Ed Solomon", "country of citizenship", "American"], ["Tran Anh Hung", "country of citizenship", "French"], ["Tran Anh Hung", "country of citizenship", "Vietnamese"]]\}} \\
\\
\textbf{\large \# Iteration 1} \\
\textbf{<Retrieval Module>} \\
{[Retrieval]} Passage 0: (Title: I Come with the Rain) I Come with the Rain I Come with the Rain is a 2009 neo-noir atmospheric thriller written and directed by Vietnamese-born French director Tran Anh Hung, ..., The film score is written \\
\\
\textbf{<Leaf Module>} \\
Leaf Instruction: Given the following Question and Passages, please distillate useful information from the Passages to address the Question effectively and list the support passage index for each distilled information. Your response should be under the format \{"useful\_information": [\{"info": statement of distilled useful information combining the question, "support\_passages": [indexes of support passages]\}]\}. Not provided information should not appear in your response. Please generate a dict format response. \\
{[Passage 0]} \\
{Question: [Original Question] Your response:}\\
{\{"useful\_information": [\{"info": "I Come with the Rain is directed by Vietnamese-born French director Tran Anh Hung.", "support\_passages": [0]\}]\}} \\
\\
\textbf{<Memory Module>} \\
Known Info: I Come with the Rain is directed by Vietnamese-born French director Tran Anh Hung. \\
Retrieved Passage: [Passage 0] \\
Historical Questions: Do both films Levity (Film) and I Come With The Rain have the directors that share the same nationality? \\
\\
\textbf{<Main Module>} \\
Main Instruction: Answer the Question based solely on the provided Information. If the Information is insufficient to answer the Question, the answer should be "unanswerable", and you should provide your explanation and a summary of missing information. Otherwise, write an accurate and concise answer to the Question with a confidence score varying from 1 (not confident) to 5 (very confident) then explain. Your response should be under the format {"answer": {"text": your answer, "confidence": confidence score}, "explanation": your explanation, "missing\_information": the summary of missing information}. Please generate a dict format response. \\
Information: [Known Info] \\
Question: [Original Question] Your response: \\
\{"answer": \{"text": "unanswerable", "confidence": 1\}, "explanation": "The information provided only specifies the nationality of Tran Anh Hung, the director of "I Come With The Rain." There is no information provided about the nationality of the director of the film "Levity." Without knowing the nationality of the director of "Levity," it is impossible to determine if both films have directors who share the same nationality.", "missing\_information": "The nationality of the director of the film "Levity.""\} \\
\\
\textbf{\large \# Iteration 2} \\
\textbf{<Retrieval Module>} \\
Query Instruction: Based on the Original Question, Historical Questions, Known Information and Missing Information, write no more than 3 queries that ask for Missing Information to solve the Original Question. If the missing information is multi-hop, decompose it into several simple and single-hop queries. The new queries should not contain redundant information and should differ from the Original Question and Historical Questions. \\
{[Original Question]} \\
{[Historical Questions]} \\
{[Known Info]} \\
Missing Information: The nationality of the director of the film "Levity. New queries: \\
What is the nationality of the director of the film "Levity"? \\
Who directed the film "Levity"? \\
Can you provide the nationality of the director of the film "Levity"? \\
\\
{[Retrieval]} Passage 1: (Title: Levity (film)) Levity (film) Levity is a 2003 drama film directed by Ed Solomon starring Billy Bob Thornton, ..., After spending his time staring at a clipping of Abner Easley, the boy he killed, he returns to the city he used to live in to find redemption. He ends up living \\
\\
\textbf{<Leaf Module>} \\
{[Leaf Instruction]} \\
{[Passage 1]} \\
{Question: Who directed the film "Levity"?} Your response: \\
{\{"useful\_information": [\{"info": "The film "Levity" was directed by Ed Solomon.", "support\_passages": [1]\}]\}} \\
\\
\textbf{<Memory Module>} \\
Known Info: I Come with the Rain is directed by Vietnamese-born French director Tran Anh Hung. The film "Levity" was directed by Ed Solomon.\\
Retrieved Passage: [Passage 0, Passage 1] \\
Historical Questions: Do both films Levity (Film) and I Come With The Rain have the directors that share the same nationality? What is the nationality of the director of the film "Levity"? Who directed the film "Levity"? Can you provide the nationality of the director of the film "Levity"?\\
\\
\textbf{<Main Module>} \\
{[Main Instruction]} \\
Information: [Known Info] \\
Question: [Original Question] Your response: \\
\{"answer": \{"text": "unanswerable", "confidence": 1\}, "explanation": "Without knowing the nationality of the director of 'Levity,' it's impossible to determine whether the directors of both films share the same nationality.", "missing\_information": "The nationality of the director of the film "Levity.""\} \\
\\
\textbf{\large \# Iteration 3} \\
\textbf{<Retrieval Module>} \\
{[Query Instruction]} \\
{[Original Question]} \\
{[Historical Questions]} \\
{[Known Info]} \\
Missing Information: The nationality of the director of the film "Levity". New queries: \\
What is the nationality of Ed Solomon, the director of the film "Levity"? \\
Can you confirm the nationality of Ed Solomon, the director of the film "Levity"? \\
\\
{[Retrieval]} Passage 2: (Title: Ed Solomon) Edward James Solomon (born September 15, 1960) is an American filmmaker. \\
\\
\textbf{<Leaf Module>} \\
{[Leaf Instruction]} \\
{[Passage 2]} \\
{Question: What is the nationality of Ed Solomon, the director of the film "Levity"?} Your response: \\
{\{"useful\_information": [\{"info": "Ed Solomon is an American filmmaker", "support\_passages": [2]\}]\}} \\
\\
\textbf{<Memory Module>} \\
Known Info: I Come with the Rain is directed by Vietnamese-born French director Tran Anh Hung. The film "Levity" was directed by Ed Solomon. Ed Solomon is an American filmmaker\\
Retrieved Passage: [Passage 0, Passage 1, Passage 2] \\
Historical Questions: Do both films Levity (Film) and I Come With The Rain have the directors that share the same nationality? What is the nationality of the director of the film "Levity"? Who directed the film "Levity"? Can you provide the nationality of the director of the film "Levity"? What is the nationality of Ed Solomon, the director of the film "Levity"? Can you confirm the nationality of Ed Solomon, the director of the film "Levity"?\\
\\
\textbf{<Main Module>} \\
{[Main Instruction]} \\
Information: [Known Info] \\
Question: [Original Question] Your response: \\
\{"answer": \{"text": "No", "confidence": 5\}, "explanation": "The director of 'Levity', Ed Solomen, is American, while the director of 'I Come With The Rain', Tran Anh Hung, is Vietnamese-born. Therefore, the directors of these films have different nationalities.", "missing\_information": ""\} \\

\\
\textbf{\large \# Case 2} \\
<Orignal Question> \\
Which film has the director died earlier, Payment On Demand or My Cousin From Warsaw? \\
<Label> \\
{\{"answer": "My Cousin From Warsaw", "evidences": [["Payment on Demand", "director", "Curtis Bernhardt"], ["My Cousin from Warsaw", "director", "Carl Boese"], ["Curtis Bernhardt", "date of death", "22 February 1981"], ["Carl Boese", "date of death", "6 July 1958"]]\}} \\
\\
\textbf{\large \# Iteration 1} \\
\textbf{<Retrieval Module>} \\
{[Retrieval]} Passage 0: My Cousin from Warsaw is a 1931 German comedy film directed by Carl Boese and starring Liane Haid, Tala Birell and Fritz Schulz. A separate French- language version" Ma cousine de Varsovie" was also made, directed by Carmine Gallone. \\
Passage 1: My Cousin from Warsaw is a 1931 French- German comedy film directed by Carmine Gallone and starring Elvire Popesco, André Roanne and Madeleine Lambert. \\
\\
\textbf{<Leaf Module>} \\
{[Leaf Instruction]} \\
{[Passage 0, Passage 1]} \\
{Question: Which film has the director died earlier, Payment On Demand or My Cousin From Warsaw?} Your response: \\
{\{"useful\_information": [\{"info": "My Cousin from Warsaw is a 1931 German comedy film directed by Carl Boese", "support\_passages": [0]\}]\}, \{"info": "My Cousin from Warsaw is a 1931 French-German comedy film directed by Carmine Gallone", "support\_passages": [1]\}]\}} \\
\\
\textbf{<Memory Module>} \\
Known Info: My Cousin from Warsaw is a 1931 German comedy film directed by Carl Boese; My Cousin from Warsaw is a 1931 French-German comedy film directed by Carmine Gallone\\
Retrieved Passage: [Passage 0, Passage 1] \\
Historical Questions: Which film has the director died earlier, Payment On Demand or My Cousin From Warsaw?\\
\\
\textbf{<Main Module>} \\
{[Main Instruction]} \\
Information: [Known Info] \\
Question: [Original Question] Your response: \\
\{"answer": \{"text": "unanswerable", "confidence": 1\}, "explanation": "The given information does not provide details about the directors of either film and their respective dates of death, making it impossible to determine which director died earlier based on the provided information.", "missing\_information": "The death dates of the directors of 'Payment On Demand' and 'My Cousin From Warsaw' are required."\} \\
\\
\textbf{\large \# Iteration 2} \\
\textbf{<Retrieval Module>} \\
{[Query Instruction]} \\
{[Original Question]} \\
{[Historical Questions]} \\
{[Known Info]} \\
Missing Information: The death dates of the directors of 'Payment On Demand' and 'My Cousin From Warsaw' are required. New queries: \\
What is the death date of the director of "Payment On Demand"? \\
When did the director of "Payment on Demand" died? \\
When did the director of My Cousin From Warsaw die? \\
\\
{[GPT knowledge prompting]} The director of the film "Payment On Demand" was Curtis Bernhardt. He was born on April 15, 1899, in Worms, Germany, and died on February 22, 1981, in Pacific Palisades, California, United States. \\
\\
\textbf{<Leaf Module>} \\
{[Leaf Instruction]} \\
{[GPT knowledge]} \\
{Question: What is the death date of the director of "Payment On Demand"?} Your response: \\
{\{"useful\_information": [\{"info": "The director of the film 'Payment On Demand,' Curtis Bernhardt, died on February 22, 1981", "support\_passages": [GPT knowledge]\}]\}} \\
\\
\textbf{<Memory Module>} \\
Known Info: My Cousin from Warsaw is a 1931 German comedy film directed by Carl Boese; My Cousin from Warsaw is a 1931 French-German comedy film directed by Carmine Gallone; The director of the film 'Payment On Demand,' Curtis Bernhardt, died on February 22, 1981\\
Retrieved Passage: [Passage 0, Passage 1, GPT knowledge] \\
Historical Questions: Which film has the director died earlier, Payment On Demand or My Cousin From Warsaw? What is the death date of the director of "Payment On Demand"? When did the director of "Payment on Demand" died? When did the director of My Cousin From Warsaw die? \\
\\
\textbf{<Main Module>} \\
{[Main Instruction]} \\
Information: [Known Info] \\
Question: [Original Question] Your response: \\
\{"answer": \{"text": "unanswerable", "confidence": 1\}, "explanation": "The provided information does not contain the death date of the director of My Cousin from Warsaw. Hence, it is impossible to determine which film's director died earlier based on the given information.", "missing\_information": "The death date of the director of My Cousin from Warsaw."\} \\
\\
\textbf{\large \# Iteration 3} \\
\textbf{<Retrieval Module>} \\
{[Query Instruction]} \\
{[Original Question]} \\
{[Historical Questions]} \\
{[Known Info]} \\
Missing Information: The death date of the director of My Cousin from Warsaw. New queries: \\
When did the director of "My Cousin From Warsaw" (directed by Carl Boese) die?\\
When did the director of "My Cousin From Warsaw" (directed by Carmine Gallone) die?
\\
\\
{[Retrieval]} Passage 2: Carmine Gallone( 10 September 1885 – 4 April 1973) was an early acclaimed Italian film director, screenwriter, and film producer. \\
Passage 3: Carl Boese( 26 August 1887 – 6 July 1958) was a German film director, screenwriter and producer. He directed 158 films between 1917 and 1957. \\
\\
\textbf{<Leaf Module>} \\
{[Leaf Instruction]} \\
{[Passage 2, Passage 3]} \\
{Question: When did the director of "My Cousin From Warsaw" (directed by Carl Boese) die? When did the director of "My Cousin From Warsaw" (directed by Carmine Gallone) die?} Your response: \\
{\{"useful\_information": [\{"info": "Carl Boese died on July 6, 1958", "support\_passages": [3]\}]\}, \{"info": "Carmine Gallone died on April 4, 1973", "support\_passages": [2]\}]\}} \\
\\
\textbf{<Memory Module>} \\
Known Info: My Cousin from Warsaw is a 1931 German comedy film directed by Carl Boese; My Cousin from Warsaw is a 1931 French-German comedy film directed by Carmine Gallone; The director of the film 'Payment On Demand,' Curtis Bernhardt, died on February 22, 1981; Carl Boese died on July 6, 1958; Carmine Gallone died on April 4, 1973\\
Retrieved Passage: [Passage 0, Passage 1, GPT knowledge, Passage 2, Passage 3] \\
Historical Questions: Which film has the director died earlier, Payment On Demand or My Cousin From Warsaw? What is the death date of the director of "Payment On Demand"? When did the director of "Payment on Demand" died? When did the director of My Cousin From Warsaw die? When did the director of "My Cousin From Warsaw" (directed by Carl Boese) die? When did the director of "My Cousin From Warsaw" (directed by Carmine Gallone) die?\\
\\
\textbf{<Main Module>} \\
{[Main Instruction]} \\
Information: [Known Info] \\
Question: [Original Question] Your response: \\
\{"answer": \{"text": "My Cousin from Warsaw", "confidence": 5\}, "explanation": "Based on the provided information, Carl Boese, the director of 'My Cousin from Warsaw,' died on July 6, 1958. Carmine Gallone, the director of 'My Cousin from Warsaw,' died on April 4, 1973. Therefore, the director of 'My Cousin from Warsaw' died earlier than the director of 'Payment On Demand.", "missing\_information": ""\} \\
\\
    \bottomrule
    \caption{Case of MIGRES. We utilize the same instruction for Multi-hop QA and ODQA. StrategyQA, however, is a categorization task which requires the model to output "yes" or "no" answer. Therefore, the instruction of StrategyQA is slightly different in the Main Module, where we add "your task is to reason from the Information whether the answer to the Question is "yes" or "no"" to make the model response as expected. We omit the full text for saving space.}
    \label{migres_case}
\end{longtable}

\end{document}